\documentclass[runningheads]{llncs}
\usepackage{graphicx}
\graphicspath{{./images/}}
\usepackage{tikz}
\usepackage{comment}
\usepackage{amsmath,amssymb} % define this before the line numbering.
\usepackage{color}
\usepackage{caption}
\usepackage{subcaption}
\usepackage{multirow}
\usepackage{adjustbox}
\usepackage[flushleft]{threeparttable}
\usepackage{booktabs}

\definecolor{blue}{RGB}{0, 0, 255}

\usepackage[accsupp]{axessibility}  % Improves PDF readability for those with disabilities.

\begin{document}
% \linenumbers
\pagestyle{headings}
\mainmatter
\def\ECCVSubNumber{3632}  % Insert your submission number here

\title{Weakly Supervised Object Localization via Transformer with Implicit Spatial Calibration}

%\title{Dynamic Diffused Module for Weakly Supervised Object Localization via Transformer} % Replace with your title

\begin{comment}
\titlerunning{ECCV-22 submission ID \ECCVSubNumber} 
\authorrunning{ECCV-22 submission ID \ECCVSubNumber} 
\author{Anonymous ECCV submission}
\institute{Paper ID \ECCVSubNumber}
\end{comment}
%******************

% \begin{comment}
\titlerunning{Transformer with Implicit Spatial Calibration}
% \author{
% Haotian Bai$^1$\inst{1}\orcidID{0000-1111-2222-3333}\ \and
% Jiong Wang\inst{1,2}\orcidID{1111-2222-3333-4444} \and
% Ruimao Zhang\inst{1,2}\orcidID{2222--3333-4444-5555} \and
% Xiang Wan\int{2}
% }
\makeatletter
% *, 2, 3, ...
\renewcommand*{\@fnsymbol}[1]{\ifcase#1\or*\else\dag\fi}
\makeatother
\author{
Haotian Bai\thanks{Research done when Haotian Bai was a Research Assistant at Shenzhen Research Institute of Big Data, The Chinese Univeristy of Hong Kong
(Shenzhen). }\index{Bai, Haotian}\orcidID{0000-0002-5693-1993} \and
Ruimao Zhang \index{Zhang, Ruimao}\thanks{Corresponding Author}\orcidID{0000-0001-9511-7532}  \and
Jiong Wang \index{Wang, Jiong}\orcidID{0000-0003-3676-2544}\and
Xiang Wan \index{Wan, Xiang}
}
\authorrunning{H. Bai. et al.}
\institute{{Shenzhen Research Institute of Big Data, The Chinese Univeristy of Hong Kong (Shenzhen), China} 
\\
\email{haotianwhite@outlook.com, \ zhangruimao@cuhk.edu.cn}}

% \end{comment}
%******************
\maketitle

\begin{abstract}

Weakly Supervised Object Localization (WSOL), which aims to localize objects by only using image-level labels, has attracted much attention because of its low annotation cost in real applications.
Recent studies leverage the advantage of self-attention in visual Transformer for long-range dependency to re-active semantic regions, aiming to avoid partial activation in traditional class activation mapping (CAM).
However, the long-range modeling in Transformer neglects the inherent spatial coherence of the object, 
and it usually diffuses the semantic-aware regions far from the object boundary, making localization results significantly larger or far smaller.
To address such an issue, we introduce a simple yet effective Spatial Calibration Module (SCM) for accurate WSOL, incorporating semantic similarities of patch tokens and their spatial relationships into a unified diffusion model.
Specifically, we introduce a learnable parameter to dynamically adjust the semantic correlations and spatial context intensities for effective information propagation.
In practice, SCM is designed as an external module of Transformer, and can be removed during inference to reduce the computation cost.
The object-sensitive localization ability is implicitly embedded into the Transformer encoder through optimization in the training phase.
It enables the generated attention maps to capture the sharper object boundaries and filter the object-irrelevant background area. 
Extensive experimental results demonstrate the effectiveness of the proposed method, which significantly outperforms its counterpart TS-CAM on both CUB-200 and ImageNet-1K benchmarks. The code is available at \underline{\textcolor{blue}{https://github.com/164140757/SCM}}.


\end{abstract}

\section{Introduction}

Weakly supervised object localization (WSOL), which learns to localize objects by only using image-level labels, has attracted much attention recently for its low annotation cost. 
The representative study of WSOL, Class Activation Map (CAM) \cite{zhou2015cnnlocalization} generates localization results using features from the last convolutional layer. %
However, the model trained for classification usually focuses on the discriminative regions, resulting insufficient activation for object localization. 
To solve such an issue, there are many CNN-based methods have been proposed in the literature, including regularization \cite{DBLP:journals/corr/abs-1804-06962,Xue_2019_ICCV,Mai_2020_CVPR,Wei2022IFIA}, adversarial training \cite{DBLP:journals/corr/abs-1908-10028,Mai_2020_CVPR,DBLP:journals/corr/abs-1804-06962}, and divergent activation \cite{singh2017hideandseek,Xue_2019_ICCV,DBLP:journals/corr/abs-1905-04899}, but the CNN’s inherent limitation of local activation dampens their performance. Although discriminative activation is optimal for minimizing image classification loss, it suffers from the inability to capture object boundaries precisely.
\setlength{\textfloatsep}{5pt plus 2pt minus 2pt}
\begin{figure}[!t]
\centering
\includegraphics[width=\textwidth]{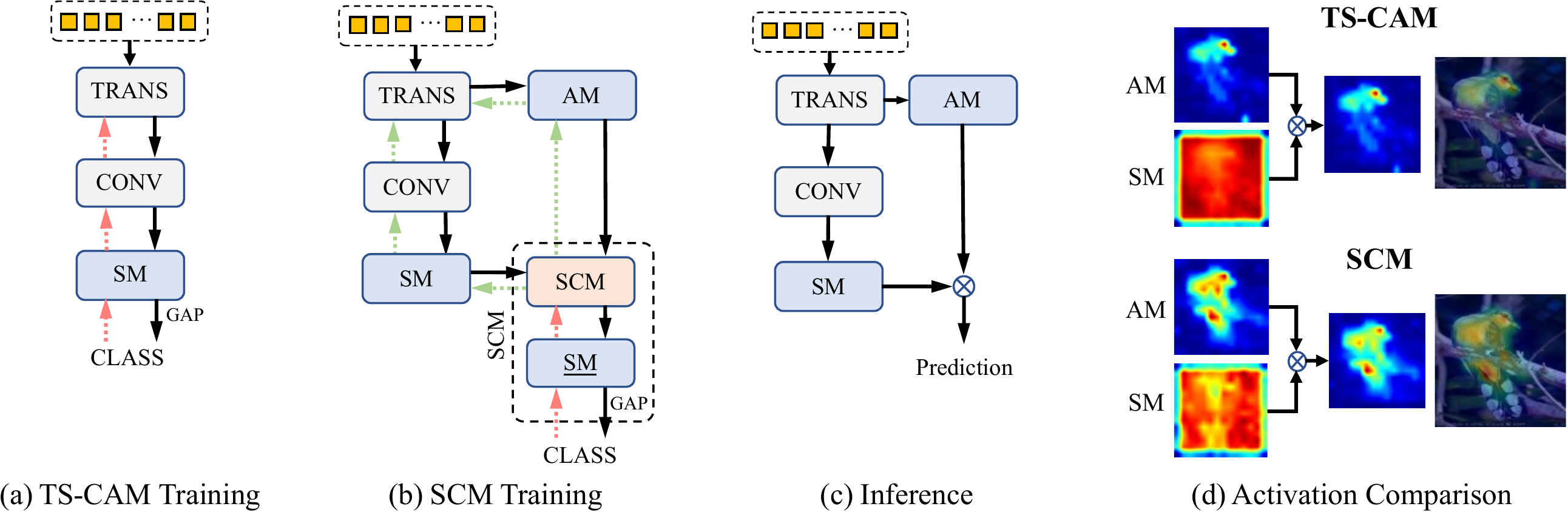}
\caption{Transformer-based localization pipelines in WSOL. The dashed arrows indicate the module parameters update during backpropagation.
(a) TS-CAM \cite{gao2021tscam}: the training pipeline encodes the feature maps into semantic maps (SM)  through a convolution head, then applies a GAP to receive gradients from the image-label supervision. 
(b) SCM(Ours): our training pipeline incorporates external SCM to produce new semantic maps \underline{SM} refined with the learned spatial and semantic correlation. Then it updates the Transformer backbone through backpropagation to obtain better attention maps and semantic representations for WOLS.
(c) Inference: SCM is dropped out, and we couple attention maps (AM) and SM just like TS-CAM for final localization prediction. 
(d) Comparison of AM, SM, and final activation maps of TS-CAM and proposed SCM.} 
\label{fig:introduction}
\end{figure}

Recently, visual Transformer has succeeded in computer vision due to its superior ability to capture long-range feature dependency. 
Vision Transformer \cite{DBLP:journals/corr/abs-2103-13915} splits an input image into patches with the positional embedding, then constructs a sequence of tokens as its visual representation. 
The self-attention mechanism enables Transformer to learn long-range semantic correlations, which is pivotal for object localization. 
A representative study is Token Semantic Coupled Attention Map (TS-CAM) \cite{gao2021tscam} which replaces traditional CNN with Transformer and takes full advantage of long-range dependencies to solve the partial activation problem. 
It localizes objects by semantic-awarded attention maps from patch tokens. 
However, we argue that only using a Transformer is not an optimal choice in practice.
Firstly, Transformer attends to long-range global dependency while inevitably it cannot capture local structure well, which is critical in describing the boundaries of objects.
In addition, Transformer splits images into discrete patches. Thus it may not attend to the inherent spatial coherence of objects, which makes it unable to predict the complete activation.
As shown in Fig.\ref{fig:introduction}(d), the activation map obtained from TS-CAM captures the global structure. Still, it concentrates in a small semantic-rich region like the bird's upper body, failing to solve partial activation completely. 
Furthermore, we observe that the fur has no abrupt change in neighboring space, and its semantic context may favor propagating the activated regions to provide a more accurate result covering the whole body.

Inspired by this potential continuity, we propose a novel external module named Spatial Calibration Module (SCM), tailored for Transformers to produce activation maps with sharper boundaries.
As shown in Fig.\ref{fig:introduction}(a)-(b), instead of directly applying  Global Average Pooling (GAP) on semantic maps to calculate loss as TS-CAM \cite{gao2021tscam}, we insert an external SCM to refine both semantic and attention maps and then use the calibrated features to calculate the semantic loss. 
Precisely, it implicitly calibrates attention representation of Transformer and produces more meaningful activation maps to cover functional areas based on spatial and contextual coherence. 
Our core design, a unified diffusion model, is introduced to incorporate semantic similarities of patch tokens and their local spatial relations during training.
While in the inference phase, SCM can be dropped out to maintain the model's simplicity, as shown in Fig.\ref{fig:introduction}(c). Then, we use the calibrated Transformer backbone to predict the localization results by coupling SM and AM. 
The main contributions of this paper are as follows: 
\begin{enumerate}
\item We propose a novel spatial calibration module (SCM) as an external Transformer module to solve the partial activation problem in WSOL by leveraging the spatial correlation. Specifically, SCM is designed to optimize Transformers implicitly and will be dropped out during inference.
\item We propose a novel information propagation methodology that provides a flexible way to integrate spatial and semantic relationships to enlarge the semantic-rich regions and cover objects completely. In practice, we introduce learnable parameters to adjust the diffusion range and filter the noise dynamically for flexible control and better adaptability. 
\item Extensive experiments demonstrate that the proposed framework outperforms its counterparts in the two challenging WSOL benchmarks. 

\end{enumerate}

\section{Related Work}
\label{sec:related}
\subsection{Weakly Supervised Object Localization.} The weakly supervised object localization aims to localize objects by solely image-level labels. The seminar work CAM \cite{zhou2015cnnlocalization} demonstrates the effectiveness of localizing objects using feature maps from CNNs trained initially for classification.
Despite its simplicity, CAM-based methods suffer from limited discriminative regions, which cannot cover objects completely. 
The field has focused on how to expand the activation with various attempts. 
Firstly, the dropout strategy is proposed to guide the model to attend to more significant regions. For instance, HaS \cite{singh2017hideandseek} hides patches in training images randomly to force the network to seek other relevant parts; CutMix \cite{DBLP:journals/corr/abs-1905-04899} adopts the same way to drop out patches but further augment the area of the patches with ground-truth labels to reduce information loss. Similarly, ADL \cite{choe2019attentionbased} adopts an importance map to maintain the informative regions' classification power. 
Instead of dropping out patches, people leverage the pixels correlations to fulfill objects as they often share similar patterns. SPG \cite{zhang2018selfproduced} learns to sense more areas with similar distribution and expand the attention scope. I$^2$C \cite{zhang2020interimage} exploits inter-and-cross images' pixel-level consistency to improve the quality of localization maps. 
Furthermore, the predicted masks can be enhanced to become complete. GC-Net \cite{lu2020geometry} highlights tight geometric shapes to fit the masks. SPOL \cite{Wei_2021_CVPR} fuses shallow features and deep features from CNN that filter the background noise and generates sharp boundaries. 

Instead of applying only CNN as the backbone for WSOL, Transformer can be another candidate to alleviate the problem of partial activation as it captures long-range feature dependency. A recent study TS-CAM \cite{gao2021tscam} utilizes attention maps from patches coupled with reallocated semantics to predict localization maps, surpassing most of its CNN counterparts in WSOL. 
Recent work LCTR \cite{chen2022lctr} adopted a similar framework with Transformer while inserting their tailored module in each Transformer block to strengthen the global features. 
However, we observe that using Transformer alone cannot solve the partial activation completely as it fails to capture the local structure and ignores spatial coherence. What is more, it is cumbersome to insert a module for each Transformer block like LCTR \cite{chen2022lctr}.
To address the issue, we propose a simple external module termed spatial calibration module (SCM) that calibrates Transformer by incorporating spatial and semantic relations to provide more complete feature maps and erase background noise.

\subsection{Graph Diffusion.} Pixels in natural images generally exhibit strong correlation, and constructing graph structure to capture such relationships has attracted much attention. In semantic segmentation, studies like \cite{DBLP:journals/corr/LiuLLLT16,DBLP:journals/corr/LiuLLLT15} build graphs on images to obtain contextual information and long-term dependencies to model label distribution jointly. 
In image preprossessing, Gene \textit{et.al} \cite{cheung2018graph} analyses graphs constructed from 2D images in spectral and succeeds in many traditional processing areas, including image compression, restoration filtering, and segmentation. The graph structure enables many classic graph algorithms and leads to new insights and understanding of image properties.  

Similarly, in WSOL, the limited activation regions share semantic coherence with neighboring locations, making it possible to expand the area by information flow to cover objects precisely. In our study, we revise the classic Graph Diffusion Kernel (GDK) algorithm \cite{kondor2002diffusion} to infer complete pseudo masks based on partial activation results. GDK is initially adopted in graph analysis like social networks \cite{bourigault2014learning}, search engines \cite{ma2011mining}, and biology \cite{Yan2008FENet} to inference pathway membership in genetic interaction networks. GDK's strategy to explore graphs via random walk inspires us to modify it to incorporate information from the image context, enabling dynamical adjustment by semantic similarity.

\section{Methodology}
This section describes the Spatial Calibration Module (SCM), which is built by stacking multiple activation diffusion blocks (ADB). ADB consists of several submodules, including semantic similarity estimation, activation diffusion, diffuse matrix approximation, and dynamic filtering. At the end of the section, we show how to predict the final localization results by using the proposed framework during the inference.

\begin{figure}[t]
\centering
\includegraphics[width=\textwidth]{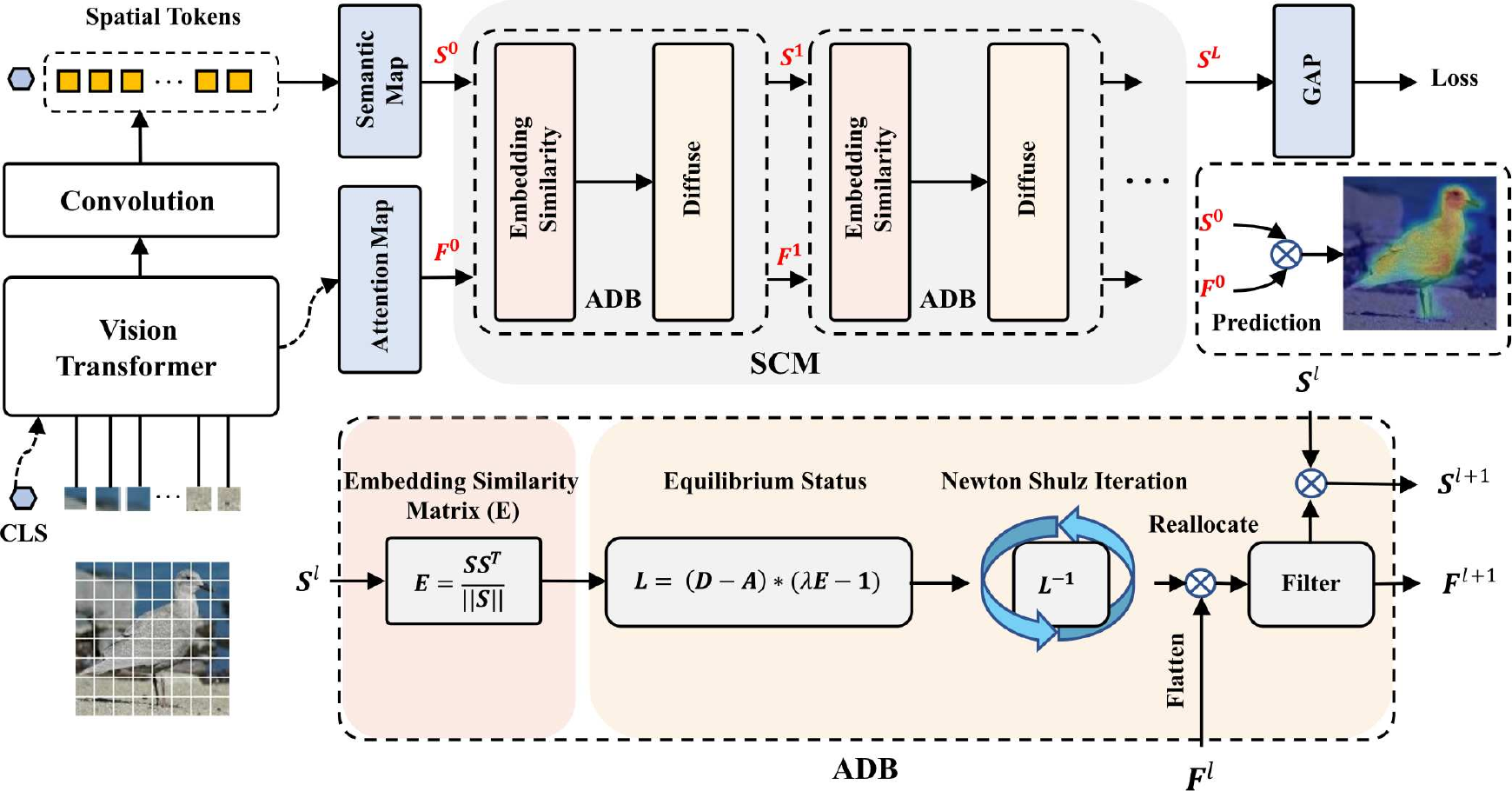}
\caption{The overall framework consists of two parts. (Left) Vision Transformer provides the original attention map $\boldsymbol{F_0}$ and semantic map $\boldsymbol{S_0}$, (Right) They are dynamically adjusted by stacked activation diffusion blocks (ADBs). The detail of the layer design is shown at the bottom-right corner (the residual connections for $\boldsymbol{F_l}$ and $\boldsymbol{S_l}$ are omitted for simplicity). Once model optimized, $\boldsymbol{F_0}$ and $\boldsymbol{S_0}$ are directly element-wise multiplied for final prediction.} 
\label{fig:arch}
\end{figure}

\subsection{Overall Architecture}
In WSOL, the attention maps from models trained on image-level labels mainly concentrate on discriminative parts, which fail to cover the whole objects.
Our proposed SCM aims to diffuse activation at small areas outwards to alleviate the partial activation problem in WSOL.
In a broad view, the whole framework is supervised by image-level labels during training. As shown in Fig.\ref{fig:introduction}(b), Transformer learns to calibrate both attention maps and semantic maps through the semantic loss from SCM implicitly. To infer the prediction, as described in Fig.\ref{fig:introduction}(c), we drop SCM and use the element-wise product of revised maps to localize objects.

As shown in Fig.\ref{fig:arch}, an input image is split into $N=H\times W$ patches with each represented as a token, where $(H, W)$ is the patch resolution.
After grouping these patch tokens and CLS token into a sequence, we send it into $I$ cascaded Transformer blocks for further representation learning. 
Similar as TS-CAM \cite{gao2021tscam}, to build the initial attention map $\boldsymbol{F}^{0} \in \mathbb{R}^{H \times W}$, 
the self-attention matrix $\boldsymbol{W}_i \in \mathbb{R}^{(N+1)\times(N+1)}$ at $i^{th}$ layer is averaged over the multiple self-attention heads.
Denote $\boldsymbol{M}_i \in  \mathbb{R}^{H\times W}$ as attention weights that corresponds to the class token in $\boldsymbol{W}_i$,
we average $\{\boldsymbol{M}_i\}_{i=1}^I$ across all intermediate layers to get the attention map $\boldsymbol{F}^{0}$ of Transformer.
%
% Then the weights are reshaped into the 2D maps with resolution ($H, W$). 
\begin{align}
{\boldsymbol{F}^0 = \frac{1}{I} \sum_{i=1}^I {\boldsymbol{M}_i}}
\end{align}
To obtain the semantic map $\boldsymbol{S}^{0} \in \mathbb{R}^{H \times W \times C}$, where $C$ denotes the number of categories, we extract all spatial tokens $ \{ \boldsymbol{t}_{n} \}_{n=1}^N$  from the last Transformer layer 
%
%where $Q$ denotes the hidden dimension of toke representation. 
%
and then encode them by a convolution head,
 \begin{align}
 \boldsymbol{S}^0 = \text{reshape} (\boldsymbol{t}_{1}... \boldsymbol{t}_{N}) * \boldsymbol{k}
 \end{align}
where $*$ is the convolution operation, $\boldsymbol{k}$ is a $3\times 3$ convolution kernel, and $\text{reshape}(\cdot)$ is an operation that converts a sequence of tokens into 2D feature maps. Then we send both  $\boldsymbol{F}^{0}$ and $\boldsymbol{S}^{0}$ into SCM to refine them. 

As illustrated in Fig.\ref{fig:arch}, for the $l^{th}$ ADB, denote ${\boldsymbol{S}}^{l}$ and $\boldsymbol{F}^{l}$ as the inputs, and $\boldsymbol{S}^{l+1}$ and $\boldsymbol{F}^{l+1}$ as the outputs. 
Firstly, to guide the propagation, we estimate embedding similarity $\boldsymbol{E}$ between pairs of patches in ${\boldsymbol{S}}^{l}$. 
To enlarge activation $\boldsymbol{F}^{l}$, we apply $\boldsymbol{E}$ to diffuse $\boldsymbol{F}^{l}$ towards the equilibrium status indicated by the inverse of Laplacian matrix $\boldsymbol{L}^{l}$. 
In practice, we re-activate $\boldsymbol{F}^{l}$ by approximating $(\boldsymbol{L}^{l})^{-1}$ with Newton Shulz Iteration. 
Afterward, a dynamic filtering module is applied to remove over-diffused parts. 
Finally, the refined $\boldsymbol{F}^{l}$ updates $\boldsymbol{S}^{l}$ via an element-wise multiplication. 

In general, by stacking multiple ADBs, the intensity of both maps is dynamically adjusted to balance semantic and spatial features.
In the training phase, we apply GAP to $\boldsymbol{S}^{L}$ to get classification logits and calculate semantic loss with the ground truth. 
During inference, SCM will be dropped out, and the element-wise product of newly extracted ${\boldsymbol{F}}^{0}$ and ${\boldsymbol{S}}^{0}$ is used to obtain the localization result.

\subsection{Activation Diffusion Block}
In this subsection, we dive into Activation Diffusion Block (ADB). Under the assumption of continuity of visual content, 
we calculate the semantic and spatial relationships of patches in $\boldsymbol{S}^{L}$, then diffuse it outwards dynamically to alleviate the partial activation problem in WSOL.

\subsubsection{Semantic Similarity Estimation.} 
Within the $l^{th}$ activation diffusion block, $l \in \{1, 2, ..., L\}$, we need semantic and spatial relationships between any pair of patches for propagation. To achieve it, we construct an undirected graph with each $\boldsymbol{v}_i^l$ connected with its first-order neighbors. Please refer to Fig.5 at the Appendix for details. Given token representation of $S^l$, we build an $N$-node graph $G^l$. 
Denote the $i^{th}$ node as $\boldsymbol{v}_i^l\in \mathbb{R}^{Q}$. Then, we can infer the semantic similarity $\boldsymbol{E}^l$, where the specific element $\boldsymbol{E}^l_{i, j}$ is defined as the cosine distance between $\boldsymbol{v}_i^l$ and $\boldsymbol{v}_j^l$:
%  whose entry $\boldsymbol{E}^l_{i, j}$ is the cosine similarity between $v_i^l$ and $v_j^l$, is defined for semantic similarity.
%
 \begin{align}
 \boldsymbol{E}^l_{i, j} = \frac{{\boldsymbol{v}}_i^l({\boldsymbol{v}}_j^l)^{\intercal}}{|| {\boldsymbol{v}_i}^l || ||{\boldsymbol{v}_j}^l||}
 \end{align}
 where $\boldsymbol{v}_i^l$ and $\boldsymbol{v}_j^l$ are flattened vectors, and the larger value $\boldsymbol{E}^l_{i, j}$ denotes the higher similarity shared by $\boldsymbol{v}_i^l$ and $\boldsymbol{v}_j^l$.

\begin{figure}[t]
\centering
\includegraphics[width=0.9\textwidth]{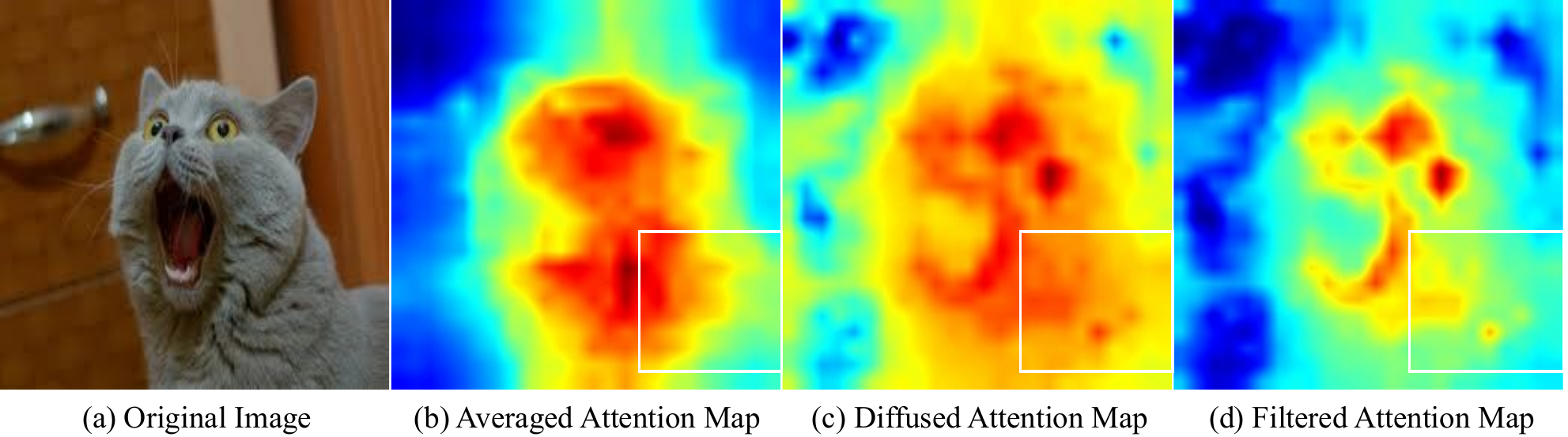}
\caption{Illustration of activation diffusion pipeline with a hand-crafted example. (a) Input image. (b) Original Transformer's attention map. (c) Diffused attention map. (d) Filtered attention map. As the spatial coherence is embedded into the attention map via our SCM, the obtained attention map by using proposed method captures a complete object boundary with less noise. } 
\label{fig:methods}
\end{figure}

\subsubsection{Activation Diffusion.} 
To present spatial relationship, we define a binary adjacency matrix $\boldsymbol{A}^l \in \mathbb{R}^{N \times N}$, whose element $\boldsymbol{A}^l_{i, j}$ indicates whether $\boldsymbol{v}_i^l$ and $\boldsymbol{v}_j^l$ are connected. We further introduce a diagonal degree matrix $\boldsymbol{D}^l \in \mathbb{R}^{N \times N}$, where $D^l_{i, i}$ corresponds to the summation of all the degrees related to $\boldsymbol{v}_i^l$. Then, we obtain Laplacian matrix $\hat{\boldsymbol{L}^l} = \boldsymbol{D}^l - \boldsymbol{A}^l$, with each element $(\boldsymbol{L}^l)^{-1}_{i, j}$ describes the correlation of $\boldsymbol{v}_i^l$ and $\boldsymbol{v}_j^l$ at the equilibrium status.

Recent studies \cite{DBLP:journals/corr/LiuLLLT16,DBLP:journals/corr/LiuLLLT15,6165311} on graph representation inspire us that the inverse of the Laplacian matrix leads to the global diffusion, which allows each unit to communicate with the rest. To enhance the diffusion with semantic relationships, we incorporate $\hat{\boldsymbol{L}^l}$ with node contextual information $\boldsymbol{E}^l$. Intuitively, we take advantage of the spatial connectivity and semantic coherence to split the tokens into the semantic-awarded foreground objects and the background environment. In practice, we use a learnable parameter $\lambda$ to dynamically adjust the semantic intensity, which makes the diffusion process more flexible and easier to fit various situations. 
The Laplacian matrix $\boldsymbol{L}^l$ with semantics is defined as,
% The shifted Laplacian matrix is used to describe coherence: 
%
 \begin{align}
 \label{eq:lap}
 {\boldsymbol{L}^l} = (\boldsymbol{D}^l - \boldsymbol{A}^l) \odot (\lambda\boldsymbol{E}^l-\boldsymbol{1})
 \end{align}
where $\odot$ represents element-wise multiplication, and $\boldsymbol{1}$ denotes the information flow exchange with neighboring vertexes. $(\boldsymbol{D}^l - \boldsymbol{A}^l)$ denotes the spatial connectivity, 
($\lambda\boldsymbol{E}^l-\boldsymbol{1}$) represents the semantic coherence, and $\odot$ incorporates them for diffusion.
Please refer to Appendix for full details of Eqn.(\ref{eq:lap}).
%
% To diffuse the original partial activation to the whole object, we need to reverse the clustering process described by $\boldsymbol{L}^l$, so we apply the inverse of ${\boldsymbol{L}^l}$ as diffuse matrix to the input attention map $\boldsymbol{F}^{l-1}$ for re-allocation of activation. 
%
After the global propagation, the reallocated activation score map can be calculated as follows,
% Then we update the the initial activation map as $\boldsymbol{F}^{l+1}$ and send it to the next layer
%
 \begin{align}
 \label{eq:UpdateF}
 \boldsymbol{F}^{l+1} = ({\boldsymbol{L}^l})^{-1} \Gamma(\boldsymbol{F}^{l})
 \end{align}
where $\boldsymbol{F}^{l+1}$ is the output re-allocated attention map and $\Gamma$ is a flattening operation that reshapes $\boldsymbol{F}^{l}$ into a patch sequence.

\subsubsection{Diffuse Matrix Approximation.} In practice, directly using $({\boldsymbol{L}^l})^{-1}$ may be impractical since ${\boldsymbol{L}^l}$ is not guaranteed to be positive-definite and its inverse may not exist. Meanwhile, as observed in our initial experiments, directly applying the inverse produced unwanted artifacts. To deal with the problems, we exploit Newton Schulz Iteration \cite{10.1145/22145.22161,10.1007/3-540-16042-6_29} to solve $({\boldsymbol{L}^l})^{-1}$ to approximate the global diffusion result,
%
%  \subsection{Optimizing}
% Equilibrium status in Eq.\ref{eq:lap} may not be the optimal for our localization task. Specifically, when the influx rate and outflux rate for $v_i$ is balanced, it doesn't mean the model redistribute activation among objects boundaries precisely, but it gives the system a thrust that drives it to diffuse outwards considering nodes' context. On the other hand, solving $(\boldsymbol{L}^l)^{-1}$ in Eq.\ref{eq:UpdateF} directly is not plausible as $\boldsymbol{L}^l$ doesn't necessarily have an inverse. Thus, we adopt Newton Schulz Iteration to approximate the equilibrium status:
%
\begin{equation} 
\begin{split}
\label{eq:NewSchu}
X_0 & = \alpha (\boldsymbol{L}^l)^{\intercal}\\
X_{p+1} & = X_{p}(2\boldsymbol{I}-\boldsymbol{L}^lX_p), %p=0, 1, ... 
\end{split}
\end{equation}
where $X_0$ is initialized as $(\boldsymbol{L}^l)^{\intercal}$ multiplied by a small constant value $\alpha$. The subscript $p$ denotes the number of iterations, and $\boldsymbol{I}$ is the identity matrix.
% As proved by Shulz\cite{schulz_1933} that eigenvalues of matrix $\boldsymbol{I}-\boldsymbol{L}^lX_p$ must be less than $1$ to assure the convergence of the scheme in Eqn.(\ref{eq:NewSchu}), which is not practical since we use image semantic data. 
% %
As discussed above, we only need $({\boldsymbol{L}^l})^{-1}$ to thrust propagation instead of obtaining the equilibrium result, so we just iterate the Eqn.(\ref{eq:NewSchu}) for $p$ times then take the approximated $({\boldsymbol{L}^l})^{-1}$ back to Eqn.(\ref{eq:UpdateF}). Then we obtain the diffused activation of $\boldsymbol{F}^{l}$, which is visualized in Fig.\ref{fig:methods}(c). 
We can see that diffusion has redistributed the averaged attention map with more boundary details, such as the \textit{ear} and the \textit{mouth}, which are beneficial for final object localization.

\subsubsection{Dynamic Filtering.} As depicted in Fig.\ref{fig:methods}(c), we found that the reallocated score map $\boldsymbol{F}^{l+1}$ provides a sharper boundary, but there is a side-effect that it diffuses the activation out of object boundaries, which may make the unnecessary background context back into $\boldsymbol{S}^{l+1}$ or result in over-estimation of bounding box. Therefore, we propose a soft-threshold filter, depicted as Eqn.(\ref{eq:dynamicfilter}), to increase density contrast between the objects and the surrounding background to depress the outside noise.

\begin{align}
\label{eq:dynamicfilter}
  \mathcal{T}(\boldsymbol{F}^{l},\beta) = \beta \cdot \text{tanhShrink}(\frac{\boldsymbol{F}^{l}}{\beta})
\end{align}
where $\beta \in (0, 1)$ is a threshold parameter for more flexible control. $\mathcal{T}$ denotes a soft-threshold function, and $\text{tanhShrink}(x) = x - \text{tanh}(x)$ is used to depress activation under $\beta$. Then $\boldsymbol{S}^{l+1}=\boldsymbol{S}^{l}\odot\mathcal{T}(\boldsymbol{F}^{l},\beta)$. As shown in Fig.\ref{fig:methods}(d), the filter operation removes noise and provides sharper contrast.

\subsection{Prediction}
After optimizing the model through backpropagation, the calibrated Transformer can generate the object-boundary-aware activation maps.
Thus, we drop SCM during inference to obtain the final bounding box. Specifically, the bounding box prediction is generated by coupling $\boldsymbol{S}^0$ and $\boldsymbol{F}^0$ as depicted in Fig.\ref{fig:arch}. As $\boldsymbol{S}^0\in \mathbb{R}^{H \times W \times C}$ is a $C$-channel 2D semantic map, each channel represents an activation map for a specific class $c$. 
To obtain the prediction from score maps, we carry out the following procedures: (1) Pass $\boldsymbol{S}^0$ through a GAP to calculate classification scores.
(2) Select $i^{th}$ map $\boldsymbol{S}_i^0\in \mathbb{R}^{H \times W}$ corresponding to the highest classification score from $\boldsymbol{S}^0$.
(3) Calculate the element-wise product $\boldsymbol{F}^0 \odot \boldsymbol{S}_i^0$. The coupled result is then up-sampled to the same size as the input for bounding box prediction.

\begin{figure}[t]
\centering
\includegraphics[width=\textwidth]{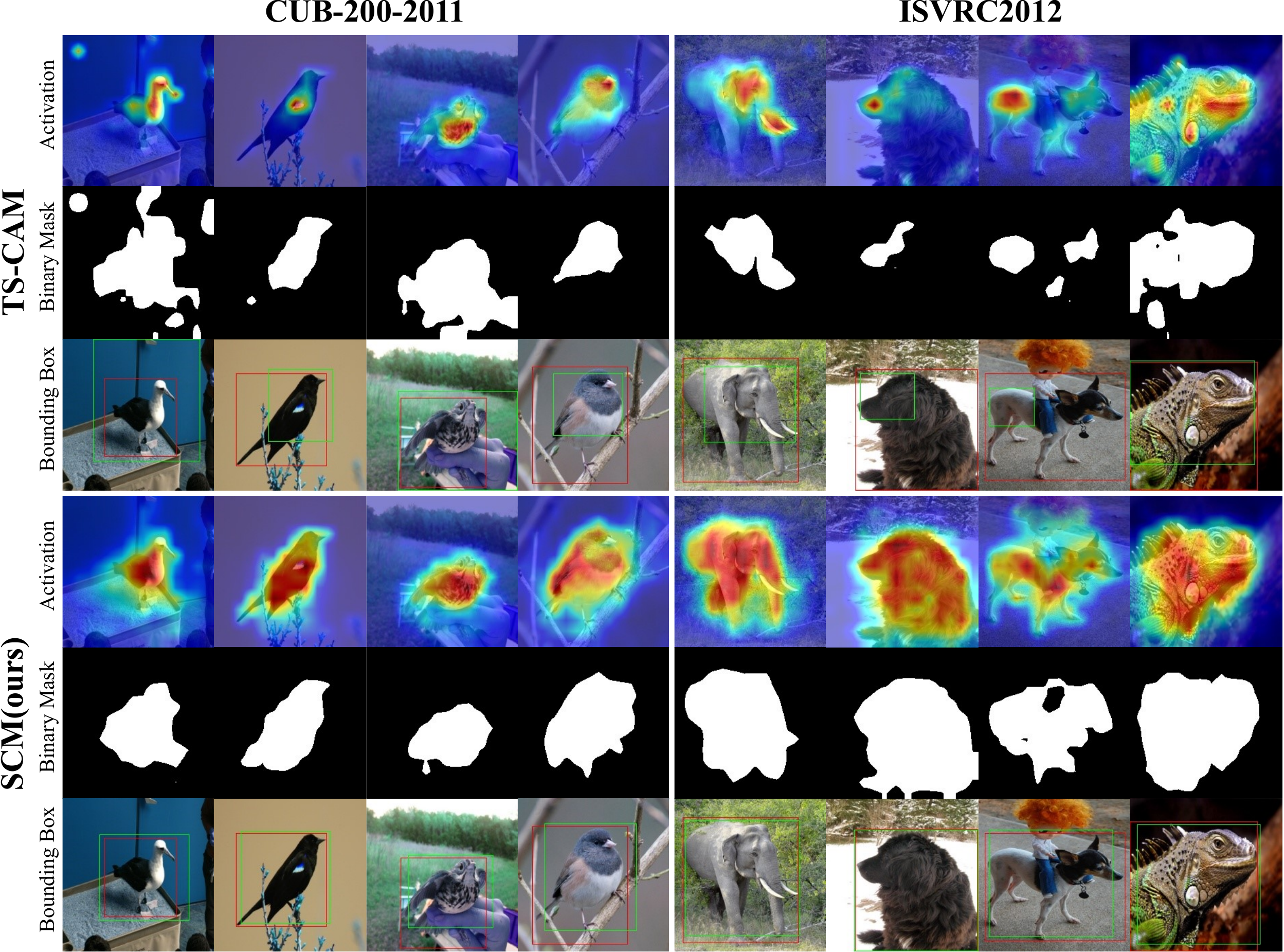}
\caption{Visual comparison of TS-CAM and SCM on 4 samples from CUB-200-2011 and ISVRC2012.
Here we use three rows for each method to show activation maps, binary map predictions, and bounding box predictions, respectively. The threshold value $\gamma$ is set to be the optimal values proposed in TS-CAM and SCM.}
\label{fig:performance}
\end{figure}

\section{Experiments}
\label{sec:experiments}
\subsection{Experiment Settings}
\subsubsection{Datasets.} We evaluate SCM on two commonly used benchmarks, CUB-200-2011 \cite{WelinderEtal2010} and ILSVRC2012 \cite{ILSVRC15}. CUB-200-2011 is an image dataset with photos of 200 bird species, containing a training set of 5,994 images and a test set of 5,794 images. ILSVRC contains about 1.2 million images with 1,000 categories for training and 50,000 images for validation. Our SCM is trained on the training set and evaluated on the validation set from which we only use the bounding box annotations for evaluation. 

\subsubsection{Evaluation Metrics.}
We evaluate the performance by the commonly used metric GT-Known and save models with the best performance. 
For GT-Known, a bounding box prediction is positive if its Intersection-over-Union (IoU) $\delta$ with at least one of the ground truth boxes is over 50\% . 
Furthermore, for a fair comparison with previous works, we apply the commonly reported Top1/5 Localization Accuracy(Loc Acc) and Classification Accuracy(Cls Acc). 
Compared with GT-Known, Loc Acc requires the correct classification result besides the condition of GT-Known.
Please refer to the appendix for more strict measures like MaxboxAccV1 and MaxboxAccV2 as recommended by \cite{choe2020evaluating} to evaluate localization performance only.

\subsubsection{Implementation details.} The Transformer module is built upon the Deit \cite{touvron2021training} pretrained on ILSVRC. In detail, we initialize $\lambda$, $\beta$ in ABDs to constant values (1 and 0.5 respectively), and choose $p=4$ and $\alpha=0.002$ in Eqn.(\ref{eq:NewSchu}). For input images, each sample is re-scaled to a size of 256$\times$256, then randomly cropped to 224$\times$224. The MLP head in the pretrained Transformer is replaced by a 2D convolution head with kernel size of 3, stride of 1, and padding of 1 to encode feature maps into semantic maps $\boldsymbol{S}^0$ (200 output units for CUB-200-2011, and 1000 for ILSVRC). The new head is initialized with He's approach \cite{DBLP:journals/corr/HeZR015}. During training, we use AdamW \cite{DBLP:journals/corr/abs-1711-05101} with $\epsilon=1e^{-8}$, $\beta_{1}=0.9$, $\beta_{2}=0.99$ and weight decay of 5e-4. On CUB-200-2011, the training lasts 30 epochs with an initial learning rate of 5e-5 and batch size of 256. On ILSVRC, the training procedure carries out 20 epochs with a learning rate of 1e-6 and batch size of 512. 
% We iteratively train the model for one epoch following a validation epoch. At last, we save the parameters with the best GT-Known performance.
We measure model performance on the validation set after every epoch. At last, we save the parameters with the best GT-Known performance on the validation set.

\subsection{Performance}
To demonstrate the effectiveness of the proposed SCM, we compare it against previous methods on CUB-200-2011 and ILSVRC2012 in Table.\ref{tb:Performance}. From GT-Known in CUB, SCM outperforms baseline method TS-CAM \cite{gao2021tscam} with a large margin, yielding GT-known 96.6$\%$ with a performance gain of 8.9$\%$. Compared with other CNN counterparts, SCM is competitive and outperforms the state-of-the-art SPOL \cite{Wei_2021_CVPR} using only about 24$\%$ parameters. As for ILSVRC, SCM surpasses TS-CAM by 1.2$\%$ on GT-Known and 5.1$\%$ on Top-1 Loc Acc and is competitive against SPOL built on the multi-stage CNN models. Compared with SPOL, SCM has the following advantages, 
(1)\textbf{Simple}: SPOL produces semantic maps and attention maps on two different modules separately, while SCM is only finetuned on a single backbone. 
(2) \textbf{Light-weighted}: SPOL is built on a multi-stage model with huge parameters, while SCM is built on a small Transformer with only about 24$\%$ parameters of the former. 
(3) \textbf{Convenient}: SPOL has to infer the prediction with the complex network design, but SCM is dropped out during the inference stage. Furthermore, compared with the recent Transformer-based works like LCTR \cite{chen2022lctr}, with the same backbone Deit-S, we surpass it by a large margin $4.2\%$ in terms of GT-Known in CUB and obtain comparable performance on Loc Acc for both CUB and ISVRC. 
We achieve this without additional parameters during inference, while other recent proposed methods add carefully designed modules or processes to improve the performance. 
The models are saved with the best GT-Known performance and achieve satisfactory Loc Acc and Cls Acc. Please refer to Sec.\ref{sec:trade-off} for more details.

The visual comparison of SCM and TS-CAM is shown in Fig.\ref{fig:performance}. We observe that TS-CAM preserves the global structure but still suffers from the partial activation problem that degrades its localization ability. 
Specifically, it cannot predict a complete component from the activation map. We notice that minor and sporadic artifacts appear on the binary threshold maps, and most of them include half parts of the objects. After adding SCM as a simple external adaptor, the masks become integral and accurate, so we believe that SCM is necessary for Transformers to find their niche in WSOL.

\begin{table*}[t]
% \footnotesize
\centering
\scalebox{0.6}{
\begin{threeparttable}
\caption{Comparison of SCM with state-of-the-art methods in both classification and localization on CUB \cite{WelinderEtal2010} and ILSVRC \cite{ILSVRC15} test set. The column Params indicates the number of parameters in backbone on which models are built. \textcolor{blue}{Values in bracket} show improvement of our method compared with TS-CAM \cite{gao2021tscam}. GT-K. stands for ground truth known.}

\label{tb:Performance}
\begin{tabular}{c|c|c|cc|ccc|cc|ccc} 
% \hline
\toprule[1.5pt]
\begin{tabular}[c]{@{}c@{}}\multirow{3}{*}{Model}\\\end{tabular} & \multirow{3}{*}{Backbone} & \multirow{3}{*}{Params(M)} & \multicolumn{5}{c|}{CUB} & \multicolumn{5}{c}{ILSVRC} \\ \cline{4-13}
& & & \multicolumn{2}{c|}{Cls Acc} & \multicolumn{3}{c|}{Loc Acc} & \multicolumn{2}{c|}{Cls Acc} & \multicolumn{3}{c}{Loc Acc} \\ \cline{4-13} 

& & & Top-1 & Top-5 & Top-1 & Top-5 & GT-K. & Top-1 & Top-5 & Top-1 & Top-5 & GT-K. \\
\hline
CAM\cite{zhou2015cnnlocalization}               & VGG16       & 138 &  -     &  -     & 34.4  & -  & - &68.8       &88.6  & 42.8 & 54.9 & - \\
ACoL\cite{DBLP:journals/corr/abs-1804-06962}    & VGG16       & 138 & 71.9  & -     & 45.9  & 61.0  & - &67.5       &88.0  & 45.8 & 63.3 &- \\
MEIL\cite{Mai_2020_CVPR}                        & VGG16       & 138 &74.8       &-       &57.5       &-       & - &73.3       &-  &49.5 &- &-  \\
% ADL\cite{choe2019attentionbased}                & VGG16       & 138 & 75.4  &       &       &       & 75.4 \\
% DDT\cite{DBLP:journals/corr/WeiZWSZ17}          & VGG16       & 138 & 84.6  &       &       &       & 84.6 \\
% RCAM\cite{DBLP:journals/corr/abs-2006-05220}    & VGG16       & 138 & 76.3  &       &       &       & 76.3 \\
SPG\cite{zhang2018selfproduced}                 & InceptionV3 & 24  &      & -      & 46.6      &59.4       &  - &\textbf{84.5} &\textbf{97.3}   &\textbf{56.1} &\textbf{70.6} &64.7 \\
I\textsuperscript{2}C\cite{zhang2020interimage} & InceptionV3 & 24  & - & -       &55.9       &68.3 &\textbf{72.6} &  73.3  &91.6       &53.1       &64.1       & 68.5 \\
GC-Net\cite{lu2020geometry}                     & GoogleNet   & 6.8     & 76.8  &\textbf{92.3}      &\textbf{63.2}       &\textbf{75.5}      & - & 77.4  &93.6  &49.1 &58.1 & -\\
ADL\cite{choe2019attentionbased}                & ResNet50-SE & 28      &80.3   &-  & 62.3  & -     & -     & 75.9  & -    & 48.5 & - & -\\
BGC\cite{kim2022bridging}                       & ResNet50    &25.6     &-      &-  & 53.8  & 65.8  & 69.9  & -     & -     & 53.8 & 65.8 & \textbf{69.9}\\
PDM\cite{meng2022pdm}                           & ResNet50    &25.6     &\textbf{81.3 }      &-       & 54.4  & 65.5  & 69.6 &75.6 &91.6 & 54.4 & 65.5 & 69.6 \\
\hline
LCTR\cite{chen2022lctr}                 & Deit-S             &22     &\textbf{85.0}       &\textbf{97.1}       &\textbf{79.2}       &89.9       &92.4 & \textbf{77.1} & \textbf{93.4} & \textbf{56.1} & 65.8 & 68.7     \\
TS-CAM\cite{gao2021tscam}                       & Deit-S      & 22  & 80.3  & 94.8  & 71.3 & 83.8 & 87.7  & 74.3 & 82.1 & 53.4 & 64.3 & 67.6\\
SCM(ours)                                       & Deit-S      & 22  &78.5 &94.5 &  76.4(\textcolor{blue}{5.1$\uparrow$}) & \textbf{91.6}(\textcolor{blue}{7.8$\uparrow$}) & \textbf{96.6}(\textcolor{blue}{8.9$\uparrow$}) &76.7(\textcolor{blue}{2.4$\uparrow$}) &93.0(\textcolor{blue}{10.9$\uparrow$}) &\textbf{56.1}(\textcolor{blue}{2.7$\uparrow$}) & \textbf{66.4}(\textcolor{blue}{2.1$\uparrow$}) & \textbf{68.8}(\textcolor{blue}{1.2$\uparrow$})\\
\hline
\multirow{2}{*}{PSOL\cite{DBLP:journals/corr/abs-2002-11359}} & DenseNet161 + & \multirow{2}{*}{95.0} &\multirow{2}{*}{-}  &\multirow{2}{*}{-} & \multirow{2}{*}{77.4}  &\multirow{2}{*}{89.5} &\multirow{2}{*}{93.0} & \multirow{2}{*}{-} & \multirow{2}{*}{-} & \multirow{2}{*}{56.4} & \multirow{2}{*}{66.5} & \multirow{2}{*}{\textbf{69.0}} \\
& EfficientNet-B7 & & & & & & & & & & & \\ 
\multirow{2}{*}{SPOL\cite{DBLP:journals/corr/abs-2002-11359}} & ResNet50 + & \multirow{2}{*}{91.6} &\multirow{2}{*}{-}  &\multirow{2}{*}{-} & \multirow{2}{*}{\textbf{80.1}}  &\multirow{2}{*}{\textbf{93.4}} &\multirow{2}{*}{\textbf{96.5}} & \multirow{2}{*}{-} & \multirow{2}{*}{-} & \multirow{2}{*}{\textbf{59.1}} & \multirow{2}{*}{\textbf{67.2}} & \multirow{2}{*}{\textbf{69.0}} \\
& EfficientNet-B7 & & & & & & & & & & & \\ 

\bottomrule[1.5pt]
\end{tabular}

\begin{tablenotes}
% \small
\item * CNN-based models are listed above. Transformer-based models are given at the center. Both PSOL \cite{DBLP:journals/corr/abs-2002-11359} and SPOL \cite{Wei_2021_CVPR} are composed of multiple-stage models are listed below. The best performance is shown as \textbf{bold} for CNN-based, Transformer, and multi-stage models, respectively.
\end{tablenotes}
\end{threeparttable}
}
\end{table*}

% \begin{table}[t]
% \caption{The comparison between SCM on different Transformers and various scales. We record GT-known and the epoch number at which the best GT-known performance is obtained.}
% \centering
% \small
% \begin{tabular}[t]{c|c|c}
% % \hline
% \toprule[1.5pt]
% Model           & Opt\_epoch  & GT-Known \\ \hline
% conformer-small & 5             & 96.1     \\
% vit-small       & 3              & 91.0     \\ \hline
% deit-small      & 20              & 96.8     \\
% deit-tiny       & 22             & 91.8     \\
% deit-base       & 5             & 93.8    \\ 
% % \hline
% \bottomrule[1.5pt]
% \end{tabular}
% \label{tb:abla_more_models}
% \end{table}

\subsection{Ablation Study}

In this section, we first illustrate the trade-off between localization and classification given the pre-determined backbone. Then we explore why SCM can reallocate and enlarge activation from two perspectives. Specifically, we show the visual results of both semantic maps $\boldsymbol{S}^l$ and attention maps $\boldsymbol{F}^l$ across all layers, and analyze them with the learnable parameters' trend during training. 
Next, we illustrate the influence of module scale by stacking a different number of ADBs. At last, we apply SCM to other Transformers like ViT \cite{DBLP:journals/corr/abs-2103-13915}, and Conformer \cite{gulati2020conformer} to prove SCM's adaptability. If not mentioned specifically, We carry out all the experiments on Deit-small with SCM consisting of four ADBs and all the experiments share the same implementation discussed above.

\begin{figure}[t]
    \centering
    \includegraphics[width=\textwidth]{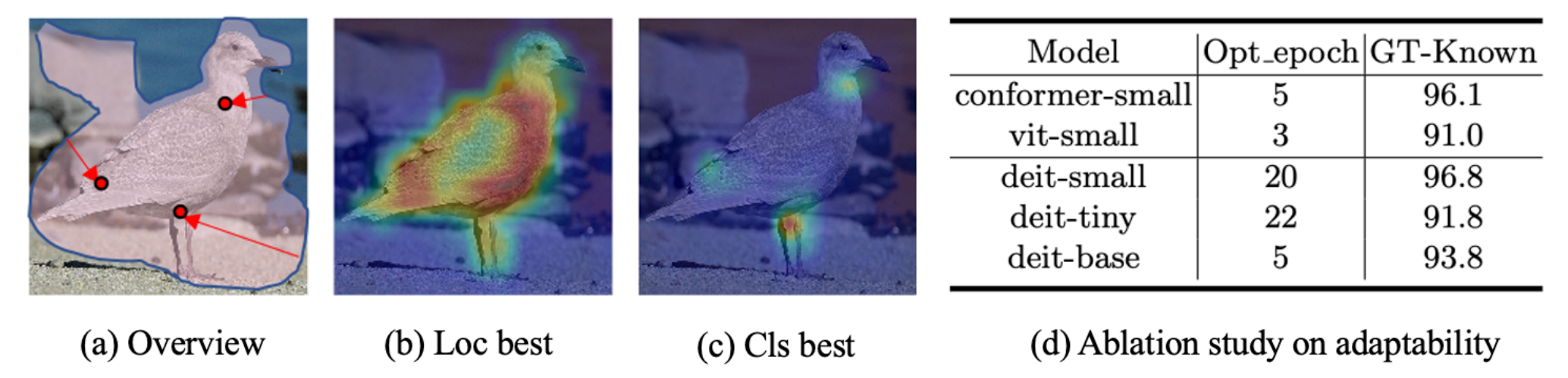}
    \caption{(a) The overview of the activation scores propagation, which is a process that evolves from the raw attention regions to the semantic rich regions. (b) Status with the best Loc Acc at the relatively early training stage. (c) Status with the best CLS Acc at the later training stage. (d) The comparison between SCM on different Transformers and various scales. We record GT-known and the epoch number at which the best GT-known performance is obtained.}
    \label{fig:trade-off}
\end{figure}

\label{sec:trade-off}
\noindent{\textbf{Trade-off between Classification \& localization. }}
SCM is an external module and will be dropped out during inference, adding no additional computational burden. 
Thus there is a trade-off between performance of localization and classification when the backbone is pre-determined.
As shown in Fig.\ref{fig:trade-off}(a), SCM aims to calibrate the raw attention to localize the bird. 
%
%As SCM is an external module, \textbf{the number of parameters is not changed}, so it is a trade-off when there are multiple goals Cls/Loc.
%
Specifically, Transformer trained with SCM localizes objects well while suffers from sub-optimal CLS Acc in Fig.\ref{fig:trade-off}(b). 
In contrast, as training process continues, it classifies objects better but only focuses on the discriminant part of the whole object, resulting in worse localization result in Fig.\ref{fig:trade-off}(c).
To clearly show the advantage of SCM for localization, we saved the model with the highest GT-Known as depicted in Fig.\ref{fig:trade-off}(b).

\begin{figure}[t]
\centering
\includegraphics[width=0.9\textwidth]{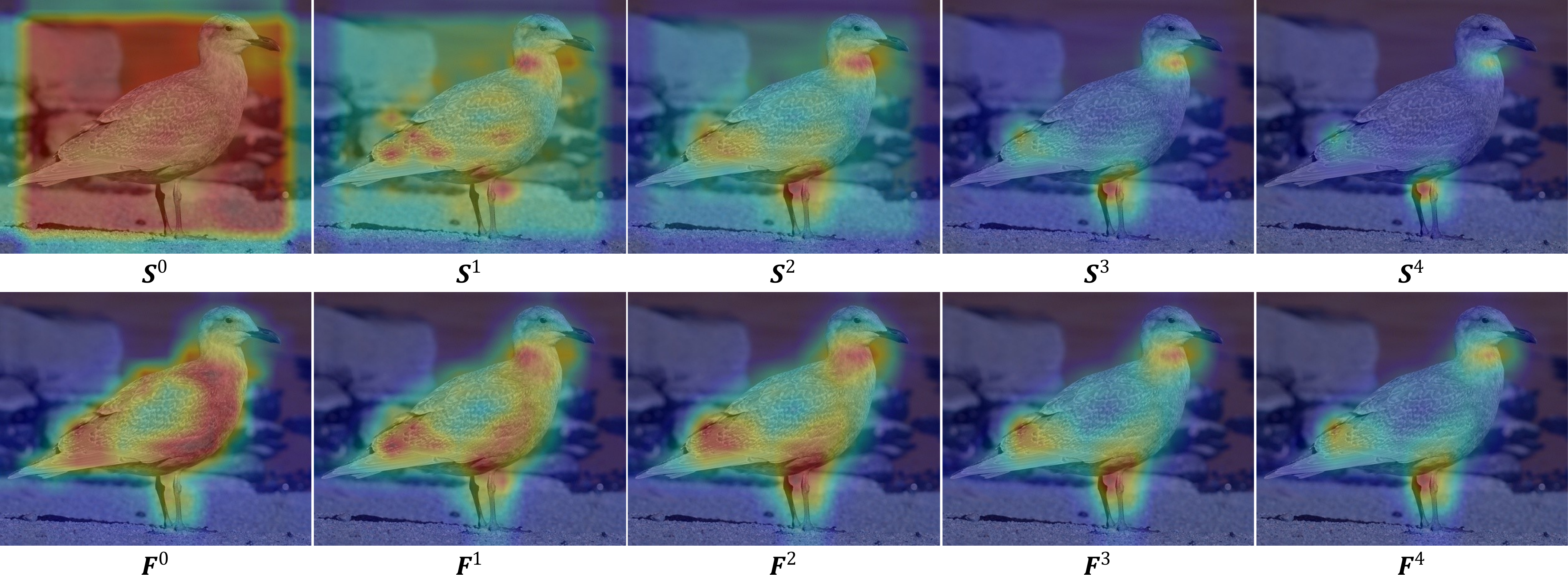}
\caption{Visualization of both semantic maps $\boldsymbol{S^l}$ (upper) and attention maps $\boldsymbol{F^l}$ (lower) input to the ${l}^{th}$ ADB block for a sample from CUB-200-2011 test set.}
\label{fig:fusion_process}
\end{figure}

\noindent{\textbf{Visualization Result of $\boldsymbol{S}^l$ and $\boldsymbol{F}^l$.}}
Implicit attention of models trained on image-level labels is blessed with remarkable localization ability as shown in CAM \cite{zhou2015cnnlocalization}. However, due to the effect of label-wise semantic loss, the models would finally be driven to gather around semantic-rich regions, causing the problem of partial activation. TS-CAM \cite{gao2021tscam} suffers from a similar issue despite improving the localization performance by Transformer's long-range feature dependency.
In Fig.\ref{fig:fusion_process}, we display both $\boldsymbol{S}^l$ and $\boldsymbol{F}^l$ at each layer of SCM. 
We observe that $\boldsymbol{F}^0$ and $\boldsymbol{S}^0$ have already covered the object completely, demonstrating that SCM can calibrate Transformer to cover objects. 
As the layer gets deeper, $\boldsymbol{S}^l$ and $\boldsymbol{F}^l$ concentrate more on semantic-rich regions, and
$\boldsymbol{S}^L$ at the last layer is further used to calculate the loss.
It explains why we drop out SCM instead of appending it to Transformer, as sharper boundaries are provided at $\boldsymbol{S}^0$ and  $\boldsymbol{F}^0$.
% In general, SCM is an external adaptor that extends the models' ability to localize objects and aims to provide sharper boundaries. 

% $\boldsymbol{S}^l$ is further to and .
% while 

\begin{figure}[t]
\centering
\includegraphics[width=\textwidth]{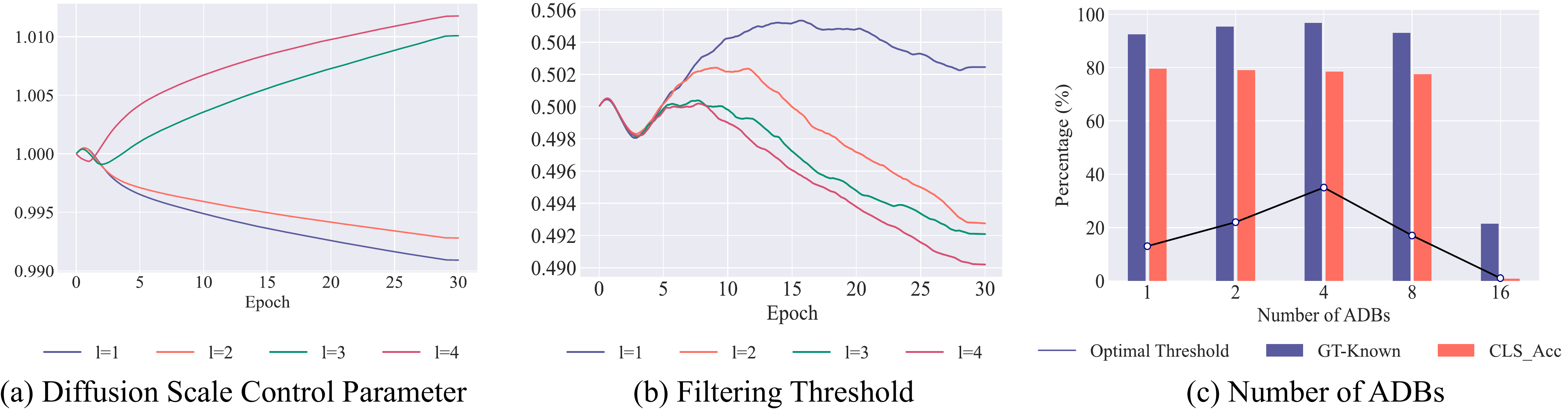}
\caption{
The learnable parameters update when trained on Deit-small. The layer number $l$ is shown below. (a) $\lambda$ is used for the diffusion scale control, and lower $\lambda$ means the wilder scale of diffusion. (b) $\beta$ determines the threshold under which the activation maps should be filtered. (c) Evaluation of GT-known, Cls Acc (top-1) for different numbers of ADBs. $\gamma$ (here in percentage format) denotes the threshold above which the bounding box is predicted from the score maps.}
\label{fig:abla_params}
\end{figure}

\noindent{\textbf{Propagating and filtering.}} To understand the effect of propagating and filtering, we analyze parameters $\lambda$ and $\beta$ in each layer of SCM. As shown in Fig.\ref{fig:abla_params}, the training record tells that $\lambda$ in deeper layers increases, while $\lambda$ in shallow layers is reduced. 
It indicates that SCM learns to diffuse activation at front layers while concentrating it in latter layers, verifying that SCM can enlarge partially activated regions with label-wise supervision. On the other hand, $\beta$ at all layers drops at the beginning, possibly because the activation provided by Transformer is sparse. It takes time for the model to shift its focus from classification to localization, as Transformer is pretrained for classification. Then it starts climbing and goes down again, indicating that attention becomes more concentrated at beginning and then turns sparse to fit the demand across layers. For instance, the front layer prefers a higher filtering threshold to reduce noise, while other layers prefer a smaller threshold to get more semantic context.

\noindent{\textbf{Stacking ADBs. }}
We further investigate the effect of module scale by stacking different numbers of ADBs. As shown in Fig.\ref{fig:abla_params}(c), we find out that the trend of GT-known and the optimal threshold almost fits the bell curve. It indicates that setting the suitable scale for SCM is essential, as when SCM becomes too deep, it fails to classify and localize objects precisely. 
% The overwhelmed diffusion may influence it since layers are more than necessary. 
On the other hand, the classification accuracy drops as the number of ADBs increases, while the localization performance increases first and drops later. It tells us that classification and localization are two different tasks, and we cannot obtain the optimal for both. 

\noindent{\textbf{Adapting SCM to more situations. }}
To evaluate SCM's performance with other Transformers, we select ViT \cite{DBLP:journals/corr/abs-2103-13915}, Conformer \cite{gulati2020conformer} to testify SCM. Next, we compare SCM on various model scales on Deit.
As shown in Fig.\ref{fig:trade-off}(d), we record the localization performance with the optimal epoch at which the best model is saved. It turns out that SCM is successfully adapted to ViT and Conformer, which achieves satisfactory performance 91.8$\%$ and 96.1$\%$ on CUB-200-2011 respectively. On the other hand, we test SCM on Deit with different scales. Surprisingly the larger models don't perform as well as Deit-small. It turns out that increasing model parameter size may not be optimal for SCM to obtain better performance, and the dropped optimal epoch number indicates that it may need a lower learning rate in training for better result. 

\noindent{\textbf{Discussions. }}
Our study presents a novel way to calibrate the Transformer for WSOL. Although we prove its adaptability to ViT \cite{DBLP:journals/corr/abs-2103-13915}, Conformer \cite{gulati2020conformer}, we cannot calibrate Transformers without CLS token such as Swin \cite{Liu2021SwinTH}, since CLS token is required to obtain $\boldsymbol{F^0}$. Furthermore, it's heuristic to choose the number of iterations used in Eqn.(\ref{eq:NewSchu}), and we simplify it as a constant number. Future research may explore methods such as Deep Reinforcement Learning to search the parameter space for the optimal diffusion policy. 
Furthermore, the equilibrium status Eqn.(\ref{eq:lap}) is a patch-wise correlation like the self-attention matrix. It may indicate a new way to find the regions of interest by diffusion.

\section{Conclusions}

We proposed a simple external spatial calibration module (SCM) to refine attention and semantic representations of Vision Transformer for weakly supervised object localization (WSOL). 
SCM exploits the spatial and semantic coherence in images and calibrates Transformers to address the issue of partial activation. To dynamically incorporate semantic similarities and local spatial relationships of patch tokens, we propose a unified diffusion model to capture sharper object boundaries and inhibit irrelevant background activation.
SCM is designed to be removed during the inference phase, and we use  Transformers' calibrated attention and semantic representations to predict localization results.
Experiments on CUB-200-2011 and ILSVRC2012 datasets prove that SCM effectively covers the full objects and significantly outperforms its counterpart TS-CAM. 
As the first Transformer external calibration module on WSOL, we hope SCM could shed light on refining Transformers for the more challenging WSOL scenarios.
\\ \hspace*{\fill} \\
\noindent{\textbf{Acknowledgement. }} The work is supported in part by the Young Scientists Fund of the National Natural Science Foundation of China under grant No. 62106154, 
by Natural Science Foundation of Guangdong Province, China (General Program) under grant No.2022A1515011524, 
by Shenzhen Science and Technology Program ZDSYS20211021111415025,
and by the Guangdong Provincial Key Laboratory of Big Data Computing, The Chinese Univeristy of Hong Kong
(Shenzhen).

%\clearpage\mbox{}Page \thepage\ of the manuscript.
%\clearpage\mbox{}Page \thepage\ of the manuscript.

%This is the last page of the manuscript.
%\par\vfill\par
%Now we have reached the maximum size of the ECCV 2022 submission (excluding references).
%References should start immediately after the main text, but can continue on p.15 if needed.
\
\clearpage
\bibliographystyle{splncs04}
\bibliography{egbib}

\clearpage

\appendix
\section{Additional ablation study}
We give more experimental results and analysis on Spatial Calibration Module (SCM) proposed in the main paper. Firstly, we conduct more ablation studies on the activation diffusion module, especially on the Newton Schulz Approximation iteration. Next, we study various strategies of combing $\boldsymbol{S}^l$ and $\boldsymbol{F}^l$ and testify its influence on localization. To test SCM on more challenging measures, we validate it on MaxboxAcc. Furthermore, we provide the complete proof of the semantics-coupled Laplacian matrix $\boldsymbol{L}^l$ at Eqn.(4) in the main paper, followed by a theoretical analysis of the semantic flow redistribution. 

In this section, we conduct experiments on the influence of the iteration number in Newton Schulz Iteration. We also test the methodology to build up the final prediction score map using maps from various layers. 

\subsection{Selecting different iteration numbers}
\begin{figure}[t]
\centering
\includegraphics[width=\textwidth]{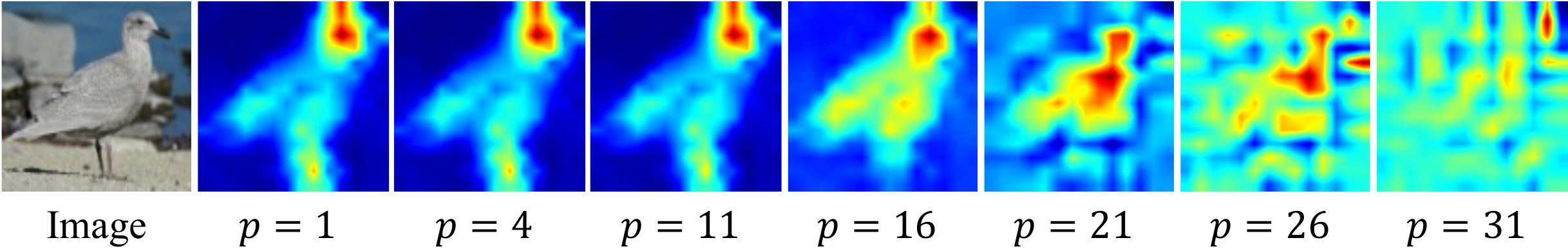}
\caption{Illustration of approximation by Newton Schulz Iteration. Each map denotes the redistributed $\boldsymbol{F}$ with corresponding number of iterations $p$ below. In the main paper, we use iteration number $p = 4$. }
\label{fig:iterations}
\end{figure}

As shown in Fig.\ref{fig:iterations}, we observe that the approximation of $\boldsymbol{L}^{-1}$ by Newton Schulz can be accelerated with the increasing number of iterations. It raises the question of what its impact on the localization performance is. To answer this question, we train several models with four ADB layers following the same setting as the main paper, except that the number of iterations varies. 

As depicted in Fig.\ref{fig:iteration_performance}, we plot GT-Known and the hyperparameter threshold $\gamma$ above which we generate the binary map. It turns out that iteration $p=4$ is still the optimal choice that exceeds other settings over 5$\%$ in GT-Known. To explore the reason, we plot $\gamma$ and observe that the iteration $p=4$ yields a much larger region of interest than others. It indicates that SCM may need a relatively small number of iterations in each block, or semantic information would be over-diffused, resulting in degraded performance.

\subsection{Strategies on combining maps}

We design SCM as an external module that calibrates Transformer trained on the classification to weakly supervised localization scenarios. During inference, SCM will be dropped out, so we only use $\boldsymbol{S}^0$ and $\boldsymbol{F}^0$ for prediction. We further explore whether combing maps from other blocks would yield a different result.

\begin{figure*}[th]
\centering
\includegraphics[width=0.6\textwidth]{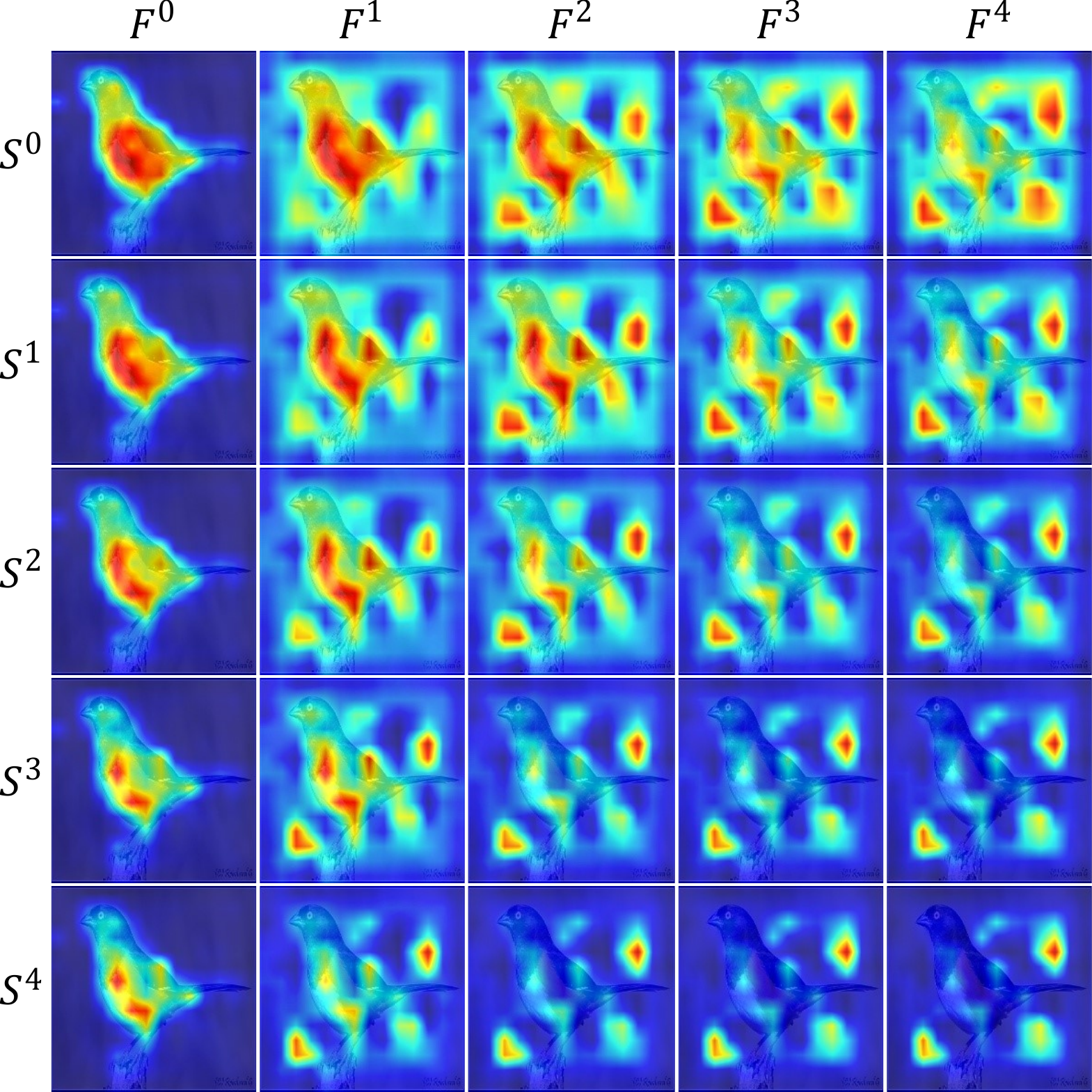}
\caption{Illustration of coupling semantic maps and attention maps across layers. Sources of each image are indicated at corresponding row and column.} 
\label{fig:SF}
\end{figure*}

As shown in Fig.\ref{fig:SF}, we produce the activation by combining $\boldsymbol{S}^l$ and $\boldsymbol{F}^l$ and depict it in a pair-wise way. We find out that $\boldsymbol{S}^l$ tends to concentrate more on semantic-rich regions as the number of layers increases. On the other hand, $\boldsymbol{F}^l$ shows a similar pattern as the layer goes deeper. The reason is that the semantic token maps $\boldsymbol{S}^l$ are supervised by the label loss that drives the model to focus on discriminative parts. However, different from the naive transformer implementation (TS-CAM), the Transformer with SCM learns to calibrate semantic and attention maps through backpropagation, as we can observe that it revises the coupled activation with more spatial details and clear boundaries in upper layers. At last, the refined coupled score map $\boldsymbol{S}^0$ and $\boldsymbol{F}^0$ becomes a promising candidate for localization.

\begin{figure*}[th]
\centering
\includegraphics[width=\textwidth]{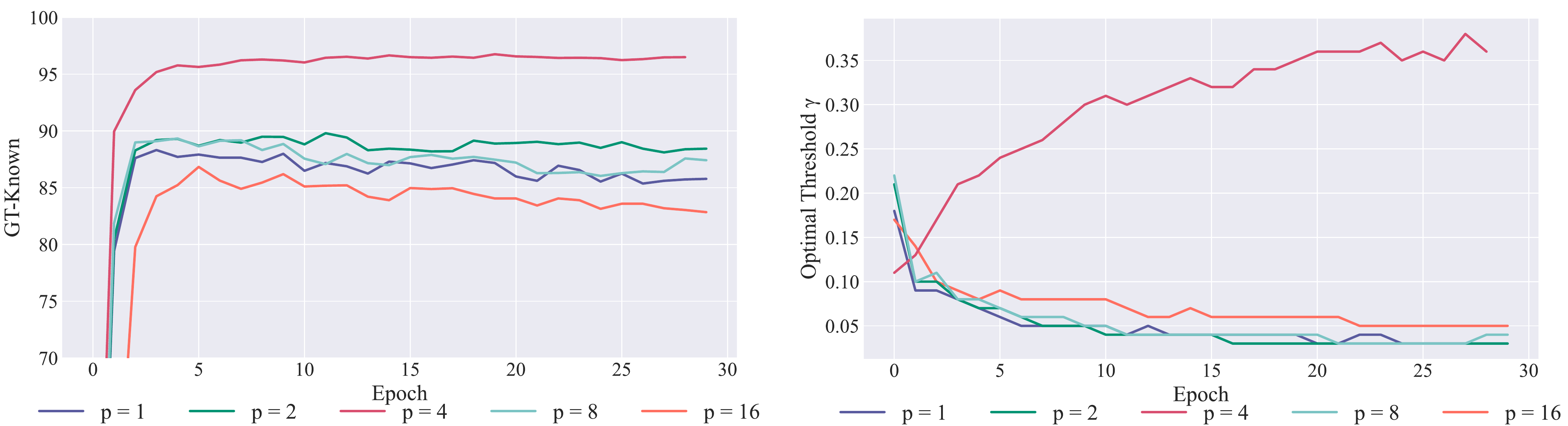}
\caption{Illustration of the GT-Known performance and the optimal filtering threshold $\gamma$ with various number of Newton Schulz Iterations $p$ in validation. $\gamma$ determines the threshold above which the bounding box is predicted from the score maps, which means $\gamma$ is proportional to the activated region.} 
\label{fig:iteration_performance}
\end{figure*}

\begin{figure*}[th]
\centering
\includegraphics[width=0.8\textwidth]{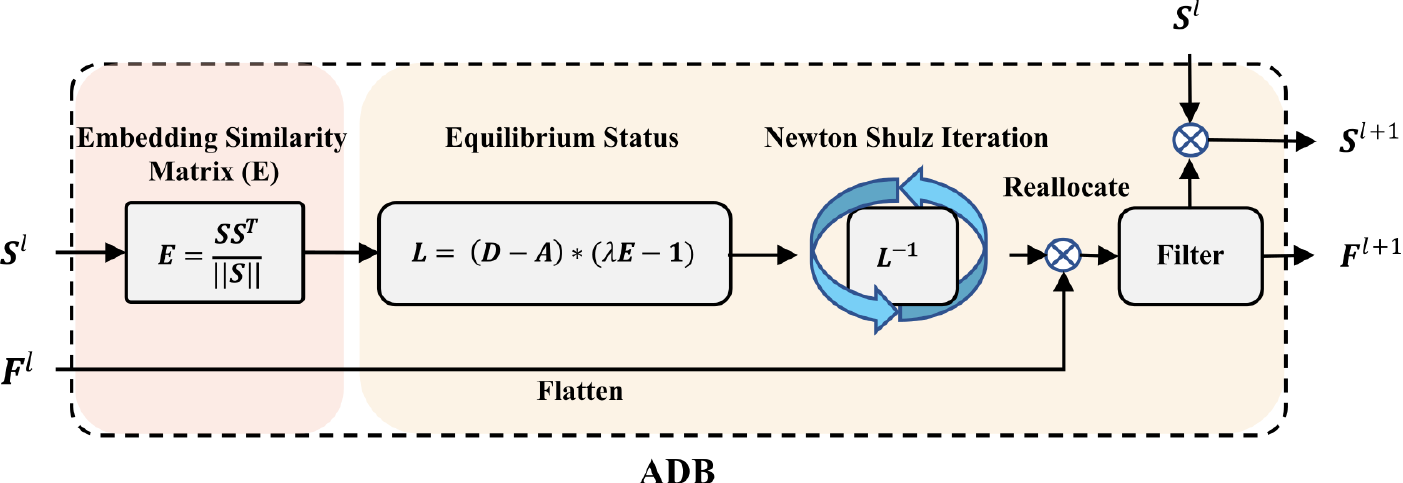}
\caption{Architecture of ${(l+1)}^{th}$ Activation Diffusion Block (ADB). }
\label{appendix:ADB_arch}
\end{figure*}

\subsection{Evaluation result on other metrics}
MaxboxAcc is reformulated to further GT-Known (the same fixed $\delta$ 50\%) with the optimal threshold $\gamma$ in generating the binary map.
 Compared with GT-Known, MaxboxAcc precludes misleading hyperparameter $\gamma$ that depends heavily on the data and model architecture. MaxboxAccV2 with optimal $\gamma$, is a more strict measure than the former MaxboxAccV1. 
 (1) It averages the performance across $\delta \in \{0.3, 0.5, 0.7\}$ to address diverse demand for localization fitness. (2) It considers the best match between the set of all estimated boxes and the set of all ground-truth boxes as prediction, instead of only one box prediction from the largest connected component of the score map in prior methods. 
 
 Towards a well-posed setup on WSOL, which is trained without any localization supervision, we shift the evaluation on a held-out set CUBV2 \cite{choe2020evaluating} not overlapping with the available validation set (now the test set). Then we evaluate both SCM and TS-CAM on it with the metrics MaxboxAccV1 and MaxboxAccV2 in the experiment shown in Table.\ref{tb:MaxboxAcc} for the reason that selecting hyperparameter $\gamma$ with full supervision in the test set violates the principle of WSOL. To make a fair comparison with the evaluation results given in \cite{choe2020evaluating}, we keep the same training budget with fixed training epochs to 50 and a fixed batch size of 32 and save the models, including SCM and TS-CAM with the best MaxboxAccV1 or MaxboxAccV2 on CUBV2.
 
 In Table.\ref{tb:MaxboxAcc}, we compare TS-CAM and SCM on CUBV2 \cite{choe2020evaluating} on the same computational budget as previous methods. It turns out that both TS-CAM and SCM have achieved satisfactory performance, but SCM surpasses TS-CAM by 7.7$\%$ and 10.3$\%$ on MaxboxAccV1 and MaxboxAccV2, respectively. Furthermore, a higher MaxboxAccV2 score proves that SCM has great adaptability and attends to various levels of localization fitness demands. 
 
\setlength{\tabcolsep}{16pt}

\begin{table}[t]
\centering
\small
\begin{threeparttable}
\caption{Comparison of SCM by MaxboxAcc \cite{choe2020evaluating} on CUB \cite{WelinderEtal2010}. Values in bracket shows improvement of our method compared with TS-CAM \cite{gao2021tscam}.}
\label{tb:MaxboxAcc}
\begin{tabular}{c|c|c|c}
% \hline
\toprule[1.5pt]
\begin{tabular}[c]{@{}c@{}}Model\\\end{tabular} & Backbone &  MaxboxAccV1 ~ & \multicolumn{1}{l}{MaxboxAccV2} \\
\hline
CAM\cite{zhou2015cnnlocalization} & VGG16 & 71.1 & 63.7 \\
ACoL\cite{DBLP:journals/corr/abs-1804-06962} & VGG16 & 72.3 & 57.4 \\
ADL\cite{choe2019attentionbased} & VGG16 & 75.7 & 66.3 \\
CutMix\cite{DBLP:journals/corr/abs-1905-04899} & VGG16 & 71.9 & 62.3 \\
SPG\cite{zhang2018selfproduced} & InceptionV3 & 62.7 & 55.9 \\
ADL\cite{choe2019attentionbased} & InceptionV3 & 63.4 & 58.8 \\ 
PDM\cite{meng2022pdm} & Resnet50 & - & 70.7\\
BGC\cite{kim2022bridging} & Resnet50 & - & 80.1\\
\hline
TS-CAM\cite{gao2021tscam} & ~ ~Deit-S &88.9  &79.6  \\
\textbf{SCM(ours)} & ~ ~\textbf{Deit-S} & \textbf{96.6} (\textcolor{blue}{7.7$\uparrow$})  & \textbf{89.9 }(\textcolor{blue}{10.3$\uparrow$})  \\
% \hline
\bottomrule[1.5pt]
\end{tabular}
\begin{tablenotes}
\small
\item * The experiment is iteratively trained one epoch on CUB train set and evaluated one epoch on CUBV2 \cite{choe2020evaluating}. The annotation mapping in the counterpart ISLVRCV2 \cite{choe2020evaluating} is currently not available, so we evaluate TS-CAM and SCM only on CUB-200-2011.
\end{tablenotes}
\end{threeparttable}
\end{table}

\section{More details on activation diffusion}
Over the past decades, Transformer has had tremendous success, largely attributed to its efficient attention mechanism to capture the long-range dependency. However, its limitations cannot be ignored. Studies have found that the transformer has a natural limitation on local context modeling \cite{DBLP:journals/corr/abs-2106-04803,DBLP:journals/corr/abs-2111-14556}, which is critical for the object localization task. We further extend its ability by introducing SCM that calibrates the Transformer to embrace spatial and semantic coherence to solve this issue. 

As shown in Fig.\ref{appendix:ADB_arch}, we apply Activation Diffusion Block in SCM to reallocate the activation region $\boldsymbol{F}$. Here, we give the detailed stpdf to get the Laplacian matrix $\boldsymbol{L}$ which denotes "equilibrium status" at Eqn.(4) in our main paper.  

\begin{figure}[t]
\centering
\includegraphics[height=5 cm]{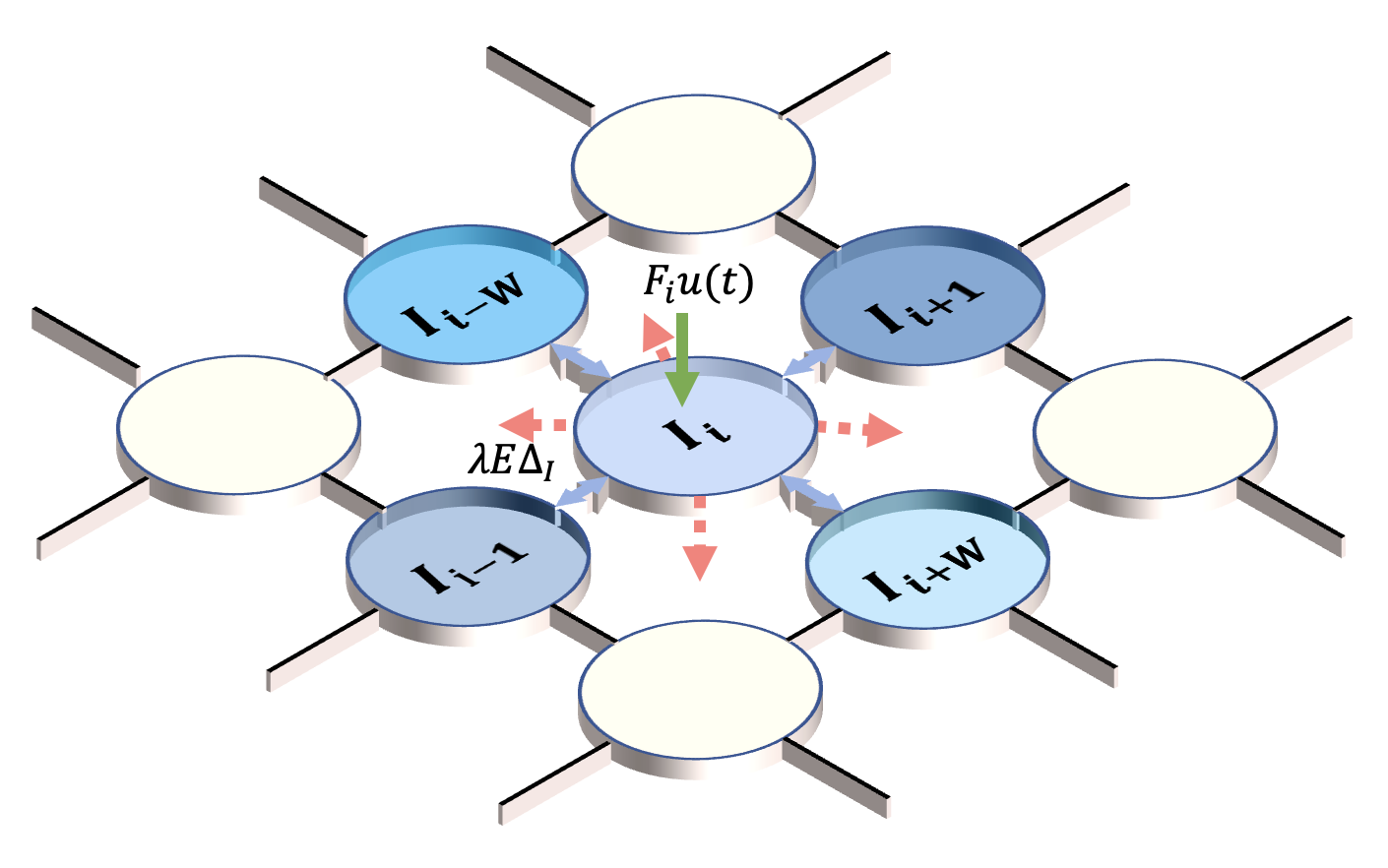}
\caption{Illustration of diffusion for $\boldsymbol{v}_i$ and its first-order neighbors, where ($H, W$) is the reshaped 2D graph resolution, where $H$ denotes the number of nodes per column, and $W$ denotes the number of nodes per row. Each circle represents a patch in this graph, and we denote the patch sequence indexes on top of them. The arrows represent flow change with the horizontal direction that denotes exchange with neighbor vertexes, and the vertical represents input and output for $G$. We further specify types of exchange by different colors, where (Green) $\boldsymbol{F}_iu(t)$ is the initial input rate; (Blue) The communication rate with neighbors; (Red) The rate of semantic flow which is related to the embedding similarity and the amount of flow.}
\label{appendix:one node in and out}
\end{figure}

This section will describe the activation diffusion behind a physic evolution model on a network structure in detail. We start with an introduction to the diffusion process that enables the exchange of information among vertexes. Next, we further analyze diffusion behavior with semantics on a global scale. At last, we show how to get the re-allocated attention map. Firstly, we build a graph $G\left \langle V, E \right \rangle$, where $V$ and $E$ represent the set of vertexes and edges, respectively. Also, $\boldsymbol{v}_i$ denotes a vertex in $V$, and $e_{i, j}$ in $E$ denotes an edge between $\boldsymbol{v}_i$ and $\boldsymbol{v}_j$, and we define the information flow as $\boldsymbol{I} \in \mathbb{R}^{N}$, where $N$ is the number of patches. $G$ is shown in Fig.\ref{appendix:one node in and out}, where we display the flow exchange between $\boldsymbol{v}_i$ and its first-order neighbors $\boldsymbol{v}_{i-1}$,  $\boldsymbol{v}_{i+1}$, $\boldsymbol{v}_{i-W}$,  $\boldsymbol{v}_{i+W}$, where $(H, W)$ is the 2D patch resolution and we use the token sequence indexes to denote the spatially connected four neighbors. 

To make diffusion semantic-aware in $G$, as shown in Fig.\ref{appendix:one node in and out}, we design a model to describe both the flow influx and the outflux on $\boldsymbol{v}_i$. Firstly, the flow input is based on the initial activation maps, where the activation score is proportional to the input rate; another source is the neighbor nodes as $\boldsymbol{v}_i$ will share flow with them. On the other hand, the flow will go outwards to nearby ones simultaneously, and to make it semantics-aware, we introduce the 'semantic flow' that escapes from the nodes. Thus, The rate of fluid change in $\boldsymbol{v}_i$ at time $t$ could be described as,
\begin{align}
\footnotesize
\hat{\boldsymbol{I}_i(t)} = \begin{matrix}\underbrace{(\boldsymbol{F}_iu(t) +  \sum_j \boldsymbol{A}_{i,j}\boldsymbol{I}_i(t)})\\influx\end{matrix} - \begin{matrix}\underbrace{(\sum_j \boldsymbol{A}_{j, i}\boldsymbol{I}_j(t) +  \lambda \sum_j \boldsymbol{A}_{i,j}(\boldsymbol{I}_i(t)-\boldsymbol{I}_j(t))\boldsymbol{E}_{i,j})}\\outflux\end{matrix}
 \label{eq:one node}
\end{align}
% \begin{align}
%  \small
% \hat{\boldsymbol{I}_i(t)} = \begin{matrix} \underbrace{ \boldsymbol{F}_iu(t) +  \sum_j \boldsymbol{A}_{i,j}\boldsymbol{I}_i(t) \\influx} \end{matrix} - \sum_j \boldsymbol{A}_{j, i}\boldsymbol{I}_j(t)   - \begin{matrix} \underbrace{ \lambda \sum_j \boldsymbol{A}_{i,j}(\boldsymbol{I}_i(t)-\boldsymbol{I}_j(t))\boldsymbol{E}_{i,j} } \\out \end{matrix}
% \label{eq:one node} 
% \end{align}
%
where $\lambda$ is a learnable parameter for flexible control over the scale of diffusion. 
Specifically, for each $\boldsymbol{v}_i$, the input for $G$ exists if $\boldsymbol{v}_i$ is one of the source nodes, then the input rate is $\boldsymbol{F}_iu(t)$, \textit{i.e.} the score maps $\boldsymbol{F}_i> 0$ then $\boldsymbol{v}_i$ can be treated as the source. Next the input from the direct neighbors should also be considered, given $\sum_j \boldsymbol{A}_{i,j}\boldsymbol{I}_i(t)$. On the other hand, for output, when propagating from $\boldsymbol{v}_i$ to $\boldsymbol{v}_j$, there exists the semantic flow which penalizes the flow exchange with low semantic similarity, \textit{i.e.} the cosine distance $\boldsymbol{E}_{i,j}$. Thus, $\lambda \sum_j \boldsymbol{A}_{i,j}(\boldsymbol{I}_i(t)-\boldsymbol{I}_j(t))\boldsymbol{E}_{i,j})$ describes the escaped semantic flows for the propagation from $\boldsymbol{v}_i$ to its neighbor, denoted as red arrows in Fig.\ref{appendix:one node in and out}. $\lambda$ is a hyperparameters to adjust the overall contribution of semantic flow. Next, the outflux into the direct neighbors is $\sum_j \boldsymbol{A}_{j, i}\boldsymbol{I}_j(t)$.

Then we can study the dynamic change of flow regarding $G\left \langle V, E \right \rangle$ and describe the graph's response to the flow dynamics. Eqn.(\ref{eq:one node}) could be further extended to the global scale, 
\begin{align}
\hat{\boldsymbol{I}(t)} =  \boldsymbol{L}\boldsymbol{I}(t) + u(t)\Gamma(\boldsymbol{F})
\label{eq:system}
\end{align}
where
\begin{align}
\boldsymbol{L} = (\boldsymbol{D}-\boldsymbol{A})*(1-\lambda \boldsymbol{E}) 
\label{ala:lap}
\end{align}
Eqn.(\ref{ala:lap}) is the shifted Laplacian matrix and $\Gamma$ is a flatten operator used to reshape $\boldsymbol{F}$ into a sequence.

Eqn.(\ref{eq:one node}) tells the flow at $\boldsymbol{v}_i$ that changes with time. Next, we could further take the integral to accumulate the total changes within a certain amount of time, which could be used to describe the trend of flow at $G$. Thus, from Eqn.(\ref{eq:system}), we obtain the expression of the amount of flow in $G$ by, 

\begin{align}
\boldsymbol{I}(t) = \int_{t^{{\prime}}=0}^{t} e^{\boldsymbol{L}(t-t^{\prime})}\Gamma(\boldsymbol{F})u(t^{\prime})dt^{\prime}
\label{eq:integral}
\end{align}

Eqn.(\ref{eq:integral}) tells us that the graph is dynamically adjusted by semantic embedding similarity $\boldsymbol{E}$ with spatial relationship. Denote a special time $t_0$ when $\hat{\boldsymbol{I}(t_0)} = 0$, we consider the 'equilibrium' status is reached as the influx rate equals the outflux rate for $\boldsymbol{v}_i$. As $t_0 \in [0,\infty]$, when $t\to \infty$, the total amount of flow in $G$ will not change and we obtain, 
\begin{align}
\label{eq:equilibrium}
  \lim\limits_{t \to \infty}\boldsymbol{I}(t) = \boldsymbol{L}^{-1}\Gamma(\boldsymbol{F})
\end{align}

Eqn.(\ref{eq:equilibrium}) implies the fully-diffused activation, however, as discussed in our main paper, $\boldsymbol{L}$ is not guaranteed to be positive-definite, and its inverse may not exist. Meanwhile, as observed in our initial experiments in Fig.\ref{fig:iterations}, directly applying the inverse has produced unwanted artifacts that may downgrade localization quality. Thanks to the Newton Schulz method, we exploit its great convergence ability that approximates $\boldsymbol{L}^{-1}$ with a few numbers of iterations. As shown in Fig.\ref{appendix:ADB_arch}, we couple the approximated $\boldsymbol{L}^{-1}$ to incorporate spatial and semantic correlation into $\boldsymbol{F}$ in the end, which is shown in Eqn.(\ref{eq:equilibrium}).
\end{document}